\documentclass[lettersize,journal]{IEEEtran}
\usepackage{amsmath,amsfonts}
\usepackage{algorithmic}
\usepackage{algorithm}
\usepackage{array}
\usepackage[caption=false,font=normalsize,labelfont=sf,textfont=sf]{subfig}
\usepackage{textcomp}
\usepackage{stfloats}
\usepackage{url}
\usepackage{verbatim}
\usepackage{graphicx}
\usepackage{hyperref}
\usepackage{cite}
\begin{document}

% \title{4DRadarRBD: 4D mmWave Radar-based Road Boundary Detection in Autonomous Driving}
\title{Road Boundary Detection Using 4D mmWave Radar for Autonomous Driving}
\author{Yuyan Wu, Hae Young Noh
        % <-this % stops a space
\thanks{Department of Civil and Environmental Engineering, Stanford University, Stanford, CA, USA (e-mail: wuyuyan@stanford.edu; noh@stanford.edu).}%
}

\markboth{}%
{Shell \MakeLowercase{\textit{et al.}}: 4DRadarRBD: 4D mmWave Radar-based Road Boundary Detection in Autonomous Driving}


\maketitle

\begin{abstract}
Detecting road boundaries, the static physical edges of the available driving area, is important for safe navigation and effective path planning in autonomous driving and advanced driver-assistance systems (ADAS). Traditionally, road boundary detection in autonomous driving relies on cameras and LiDAR. However, they are vulnerable to poor lighting conditions, such as nighttime and direct sunlight glare, or prohibitively expensive for low-end vehicles.
To this end, this paper introduces 4DRadarRBD, the first road boundary detection method based on 4D mmWave radar which is cost-effective and robust in complex driving scenarios. The main idea is that road boundaries (e.g., fences, bushes, roadblocks), reflect millimeter waves, thus generating point cloud data for the radar. 
To overcome the challenge that the 4D mmWave radar point clouds contain many noisy points, we initially reduce noisy points via physical constraints for road boundaries and then segment the road boundary points from the noisy points by incorporating a distance-based loss which penalizes for falsely detecting the points far away from the actual road boundaries.
In addition, we capture the temporal dynamics of point cloud sequences by utilizing each point's deviation from the vehicle motion-compensated road boundary detection result obtained from the previous frame, along with the spatial distribution of the point cloud for point-wise road boundary segmentation.
We evaluated 4DRadarRBD through real-world driving tests and achieved a road boundary point segmentation accuracy of 93$\%$, with a median distance error of up to 0.023 m and an error reduction of 92.6$\%$ compared to the baseline model.
\end{abstract}

\section{Introduction}

Road boundary detection is important for autonomous driving and advanced driver-assistance systems to prevent collisions. Road boundaries are the static physical edges of drivable areas, including fences, bushes, and roadblocks, beyond which there is a risk of collision. Road boundary detection helps reduce the risk of collisions and allows navigation systems to maintain a safe distance from the road boundary in autonomous driving. 

Current sensing methods for road boundary detection mainly include RGB cameras and LiDAR~\cite{sun20193d, kang2012lidar, zhang20153d, chen2017rbnet, wen2008road, taher2018proposed}. However, RGB cameras lack depth perception and struggle in poor lighting conditions. LiDAR provides accurate distance information but is costly and impractical for lower-end vehicles. Thus, autonomous driving systems require a cost-effective and robust road boundary detection method that performs reliably across various lighting conditions.

This paper introduces 4DRadarRBD, the first 4D mmWave radar-based road boundary detection system which is cost-effective and robust for complex driving scenarios. The main idea of the system is that the radar emits millimeter waves, which get reflected by road boundaries (e.g., fences, bushes, and roadblocks). These reflections are captured as point clouds, which are then used to detect the road boundaries. Unlike 3D mmWave radar, which estimates range, azimuth, and Doppler velocity, 4D mmWave radar introduces elevation measurement, significantly enhancing its ability to differentiate between various objects. The inclusion of elevation data allows the radar to more accurately identify whether an object is an overhead structure that can be safely passed beneath or a road boundary that requires maintaining a safe distance.

However, there are two main challenges for road boundary detection using 4D mmWave radar. 
Firstly, in complex driving environments, the 4D mmWave radar point cloud contains a large number of noisy points from non-road boundary objects (e.g., vehicles, overpasses) and random ghost points.
Secondly, it is difficult to capture the temporal dynamics of the point cloud sequences with the fast vehicle movement (up to 30 m/s) and the low radar sampling rate ($\sim$10 Hz). This results in a large number of points appearing at the edges of the radar's sensing range in each frame, making it challenging to maintain temporal consistency in boundary estimation.

% Brief intro of our methods
4DRadarRBD addresses these challenges and achieves road boundary detection with three modules. In the first module, the point cloud is preprocessed to extract point-wise features, reduce noise using physical constraints, and mitigate point cloud sparsity through frame fusion. In the second module, we segment the road boundary points from the noisy points and capture the temporal dynamics of the point cloud. Finally, the third module fits road boundary curves based on the segmented road boundary points, providing continuous road boundary estimates for control and planning tasks.
The first challenge is addressed by first utilizing the physical constraints of the road boundary to reduce noisy points in the first module and then incorporating a distance-based loss to penalize points far away from the road boundary for being detected as road boundaries in the second module. To solve the second challenge, we capture the temporal dynamics of point cloud sequences with the deviation of each point in the current frame from the motion-compensated road boundary points of the previous frame. The deviation is incorporated as a feature in the point cloud segmentation framework in the second module, improving temporal consistency in road boundary detection.

The main contributions of 4DRadarRBD consist of:
\begin{itemize}

\item  We introduce 4DRadarRBD, the first 4D mmWave radar-based road boundary detection system, which is cost-efficient and robust to complex driving scenarios.

\item  We mitigate the effects of noisy points and capture the temporal dynamics of point cloud sequences for robust road boundary detection.

\item  We evaluate the 4DRadarRBD system through real-world driving tests in complex scenarios and achieve accurate and robust road boundary detection results.
\end{itemize}

\section{4DRadarRBD System}
\label{sec:method}
\begin{figure}[!t]
    \centering
    \includegraphics[width=0.48\textwidth]{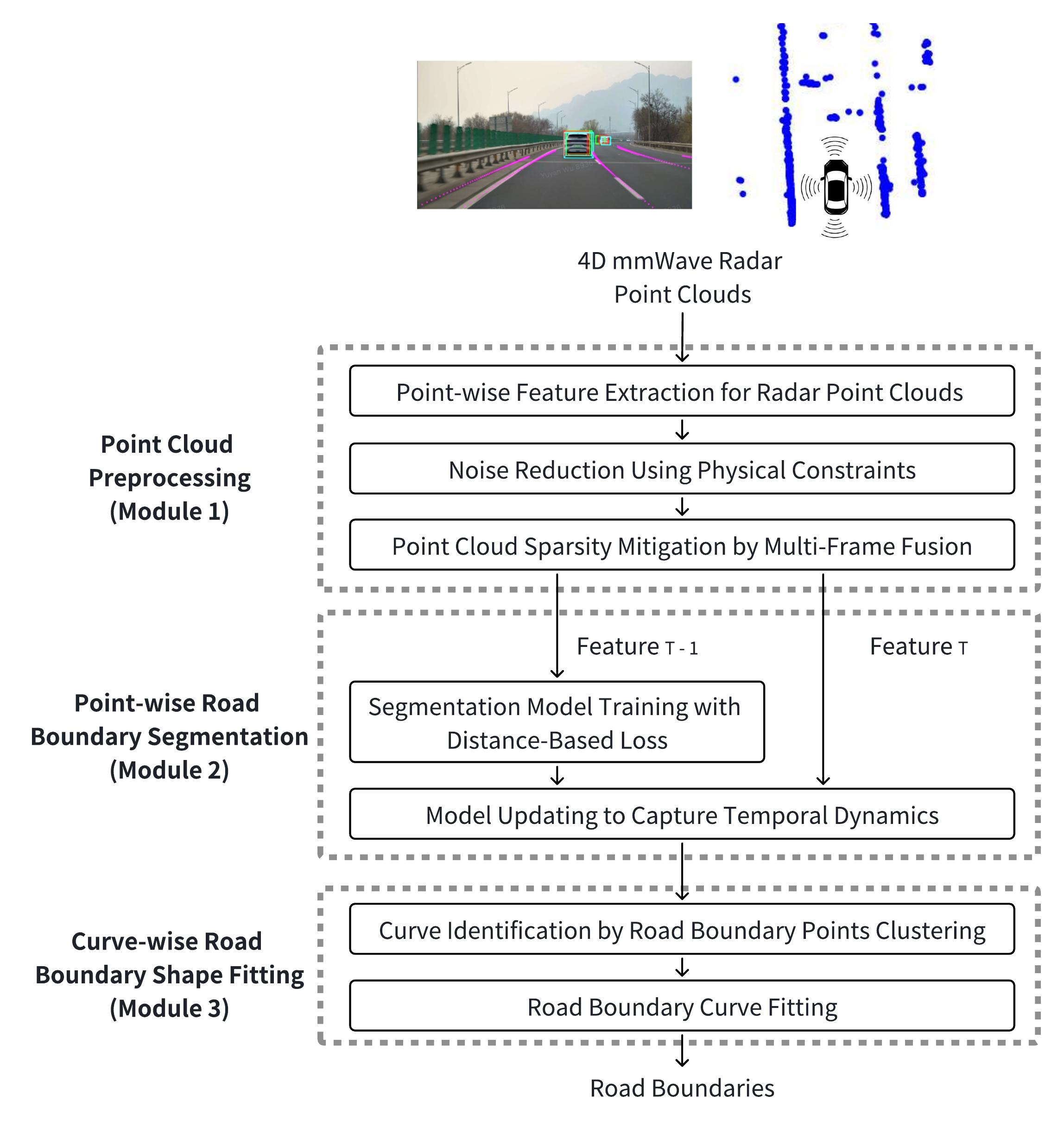}
    \caption{4DRadarRBD System Overview}
    \label{fig:system}
\end{figure}

In this section, we introduce the 4DRadarRBD system which detects road boundaries using point cloud data from 4D mmWave radar. The system mainly includes three modules: 1) point cloud preprocessing, 2) point-wise road boundary segmentation, and 3) curve-wise road boundary shape fitting  (see Fig.~\ref{fig:system}).

\subsection{Module 1: Point Cloud Preprocessing}

We first preprocess the point cloud obtained from 4D mmWave radar to extract point-wise features for the road boundary segmentation task while reducing noisy points and mitigating the sparsity of the point cloud.

% Feature Extraction for point cloud
\subsubsection{Point-wise Feature Extraction for Radar Point Clouds}

The point-wise features of the point cloud are extracted from 4D mmWave radar signals and onboard sensors (GPS, IMU), including position coordinates (x, y, z), Doppler velocity, signal-to-noise ratio, range, vehicle velocity, and yaw rate. Among them, the first four features are derived from 4D mmWave radar, while vehicle velocity and yaw rate are obtained from GPS and IMU sensors.
The position (x, y, z) is defined with respect to a coordinate system originating at the location of 4D mmWave radar, which is typically mounted near the front license plate of the vehicle. In this coordinate system, the x-axis extends to the right-hand side of the vehicle, the y-axis points forward along the vehicle’s driving direction, and the z-axis points upward.
Position and range indicate the object's location, while the signal-to-noise ratio provides insight into the material and surface properties of the object. Additionally, vehicle speed, yaw rate, and Doppler velocity collectively describe the object's motion. These features contain critical information about the object reflecting millimeter waves and are important for accurately segmenting road boundary points.

\subsubsection{Noise Reduction Using Physical Constraints}

We apply physical constraints to filter out noisy points by excluding those that exceed the expected height or velocity range of static road boundaries. Since 4DRadarRBD is designed for standard vehicles, which typically do not exceed 3 meters in height,  points above 3 meters (typically from overhead structures such as streetlights or overpasses) are excluded to reduce noise. In addition, points below -1.5 meters are likely noisy points due to elevation measurement instability or ghost points and are also excluded.
To further reduce noisy points, velocity-based filtering is applied to eliminate the points whose Doppler velocity significantly deviates from that of static road boundaries. Since road boundaries are stationary, points with substantial motion are unlikely to represent valid boundary detections. The velocity deviation is calculated as the difference between the measured Doppler velocity of a point and the expected Doppler velocity of a static object at the same position, derived from the vehicle's velocity. Points with a deviation exceeding 1 m/s are excluded as noisy points. This threshold value is selected to compensate for uncertainties in Doppler velocity and azimuth measurements.

\subsubsection{Point Cloud Sparsity Mitigation by Multi-Frame Fusion}

To mitigate the sparsity of the 4D mmWave radar point cloud, we fuse point clouds from three consecutive frames.
First, we apply motion compensation by transforming the point cloud data from previous frames into the world coordinate system and then converting it back to the self-coordinate system of the current frame. The necessary transformation and rotation matrices for this conversion are obtained from the vehicle's GPS sensor. In addition, during the fusion process, we introduce an index in the point-wise features to indicate the frame origin of each point: 0 for the current frame, 1 for the previous frame, and 2 for the frame before that. This index preserves temporal information for road boundary segmentation task.

\subsection{Module 2: Point-wise Road Boundary Segmentation}

The point-wise road boundary segmentation module aims to distinguish road boundary points from noisy points. The segmentation is based on the PointNet++ framework, which efficiently captures the spatial distribution of point clouds and point-wise features~\cite{qi2017pointnet++}. 
The main innovations of our method are: 1) we reduce the false positive detections (i.e., falsely detecting non-road boundary points) by incorporating a distance-based loss, which penalizes detected points far away from the road boundary as road boundaries into the PointNet++ segmentation network and 2) we capture the temporal dynamics of the point cloud sequences by adding the vector representing point's deviation from the motion-compensated road boundary detection result obtained from the previous frame into the point-wise features.

\subsubsection{Segmentation Model Training with Distance-Based Loss}

\begin{figure}[t]
    \centering
    \includegraphics[width=0.48\textwidth]{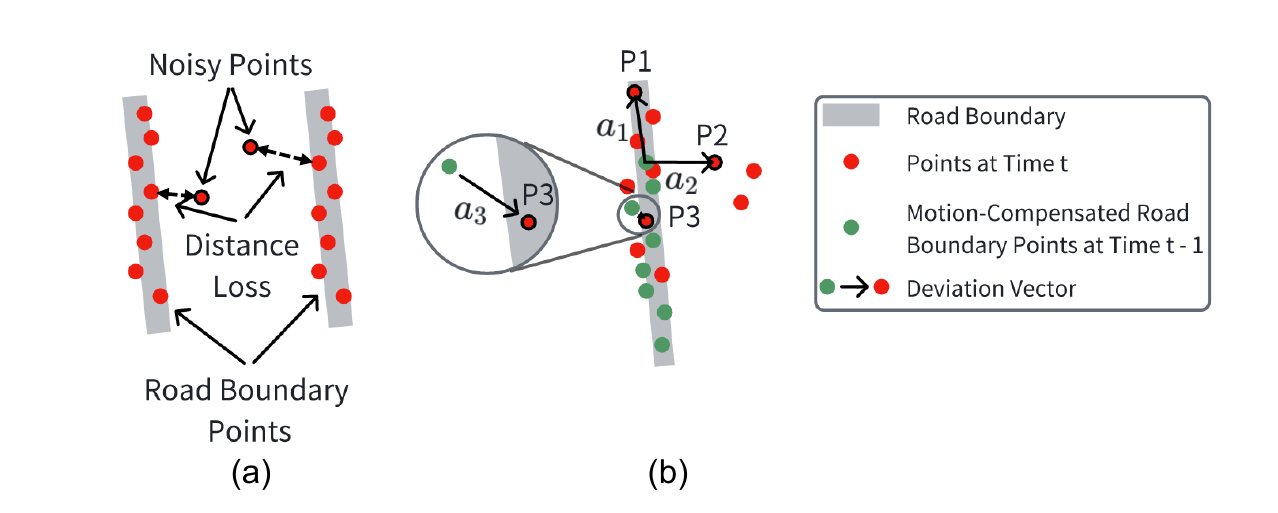}
    \caption{(a) Distance loss is calculated as the Euclidean distance between the detected and the actual road boundary points, which has a large value for the noisy points far away from the road boundaries; (b) The deviation vectors for each point in the current frame (e.g., $a_1, a_2, a_3$ for $P_1, P_2, P_3$ respectively) are calculated as the shortest vector from the motion-compensated road boundary points detected in the previous frame to the point in the current frame (e.g., P1, P2, P3). }
    \label{fig:scheme}
\end{figure}

To reduce false positive detections of noisy points, we incorporate a distance-based loss into the PointNet++ segmentation network to penalize such false positive detections. The PointNet++ network is selected as the basic structure of the point segmentation module due to its effectiveness in extracting hierarchical features and its adaptability to various spatial scales~\cite{qi2017pointnet++}.
The original PointNet++ network uses only binary cross-entropy loss for point segmentation, which is inadequate for our task. 
This limitation arises because, after point cloud preprocessing, most noisy points are filtered out, leaving only a small fraction of distant noisy points compared to road boundary points. Consequently, their influence on the loss function is minimal. However, if these distant noisy points are misclassified as road boundary points (false positive detections), they can significantly degrade the boundary fitting process.
To this end, we calculate the Euclidean distance between each detected road boundary point and its nearest actual road boundary point as the distance loss (see Fig.~\ref{fig:scheme} (a)). The distance loss is particularly high for these false positive points, which helps mitigate the false positive detections.

\subsubsection{Model Updating to Capture Temporal Dynamics}
\label{sec:update}

% Explain the challenge we are faced with
To capture the temporal dynamics of the point cloud sequences, we augment each point’s feature representation with a deviation vector, which encodes its direction and distance relative to the road boundary points detected in the previous frame.
For each point in the current frame, this deviation vector is defined as the shortest vector originating from a motion-compensated road boundary point in the previous frame and terminating at the current point (see Fig.~\ref{fig:scheme}(b)).
P1, P2, and P3, along with their respective deviation vectors $a_1, a_2, a_3$, illustrate three typical cases in the point cloud data:
(1) newly observed road boundary points (e.g., P1, $a_1$) that appear in the current frame but were absent in the previous frame due to the vehicle's forward motion, (2) noisy points that only appear in the current frame (e.g., P2, $a_2$), and (3) consistent road boundary points that appear in both the current and previous frames (e.g., P3, $a_3$). The deviation vectors exhibit different patterns for each of these three point types.
For newly appearing points (e.g., P1), the deviation vector typically points in the direction of the road boundary line. For noisy points (e.g., P2), the deviation vector usually points perpendicular to the general road boundary direction. For the nearby road boundary points (e.g., P3), the vector is usually small in magnitude since the road boundary tends to be static and continuous.
These deviation vectors ($a_1, a_2, a_3$ in Fig.~\ref{fig:scheme} (b)) provide crucial spatial and temporal information, aiding in consistent road boundary segmentation. 
Existing point-based methods for processing point cloud sequences often capture temporal dynamics by grouping points or aggregating information from neighboring points to track movements across frames~\cite{liu2019meteornet,wei2022spatial, fan2021point}. However, in autonomous driving scenarios focused on road boundary detection, temporal changes in point clouds are primarily due to the continuous appearance of new points at the edge of the sensor's range as the vehicle moves forward, rather than the movement of individual points, since road boundaries are static. Consequently, these methods may struggle to effectively handle newly appeared points, making them less suitable for this specific application.

% Explain why the probability is also added to the point-wise feature representation.
Additionally, we incorporate the road boundary probability of the closest detected road boundary point from the previous frame (the starting point of the deviation vector) as a point-wise feature. This probability serves as a confidence measure in road boundary segmentation process. Without this confidence information, if a noisy point far away from the road boundaries is mistakenly detected as a road boundary point in one frame, it can lead to error propagation in subsequent frames. However, we observe that in such cases, incorrectly detected points generally have a lower road boundary probability than actual boundary points, indicating lower confidence in the segmentation result. By incorporating this probability, the model gains confidence awareness, effectively suppressing the propagation of false detections across frames.

\subsection{Module 3: Curve-wise Road Boundary Shape Fitting}

Point-wise road boundary segmentation only provides discrete points representing road boundaries. However, control and planning tasks in autonomous driving usually require continuous road boundary curves.
To this end, we first identify the continuous curves from the detected road boundary points by clustering, and then fit the road boundary curves for each of the identified point cloud clusters.

% clustering
\subsubsection{Curve Identification by Road Boundary Points Clustering}
We employ the DBSCAN clustering method \cite{ester1996density} to identify continuous road boundary curves from the detected road boundary points. DBSCAN is chosen because it does not require a predefined number of clusters, which is suitable for situations where the number of road boundaries is uncertain. It is also capable of clustering road boundaries of various shapes, which is crucial in complex driving scenarios.
Due to the short Euclidean distance between the left and right road boundaries, the algorithm often incorrectly clusters them into one cluster. To solve this problem, we divide the y-coordinate (representing the forward direction of vehicle motion) of the point cloud by a factor before clustering (which we set to 5 in our experiments based on the observed sparsity of the point cloud). Since road boundaries usually follow the direction of vehicle motion, this scaling helps the algorithm better cluster the road boundary curves. Without this scaling, the algorithm may incorrectly cluster left- and right-parallel road boundary curves into a single cluster due to their close Euclidean distance or split a single road boundary into multiple clusters due to the variation in the distance along the y-axis.

% GPR fitting, explain the parameters chosen, the tricks for the intersection and nearby road boundaries.
\subsubsection{Road Boundary Curve Fitting}
We take subsamples from each cluster identified by the DBSCAN algorithm and use Gaussian Process Regression (GPR) to fit road boundary curves and provide 95$\%$ confidence intervals for each curve~\cite{williams1995gaussian}. 
The subsampling step aims to reduce the GPR fitting time. GPR is a nonparametric method that does not assume a specific functional form and is therefore suitable for modeling road boundaries with various shapes. We use a Matérn kernel with a $\nu$ value of 10 to avoid fitting unrealistic road boundary curves with large gradients. To prevent misconnections at intersections, where two boundary curves on opposite sides of an intersection are mistaken for a single cluster, we introduce a condition that if the point cloud data is missing along the y-axis (the forward direction of vehicle movement) for a gap of more than 6 meters (approximately the width of an urban intersection), the fitted curves will be split into two segments. In addition, if the 95\% confidence interval exceeds 2 meters, indicating uncertain identification of road boundary curves, we re-cluster the points within this cluster and fit each new cluster independently. This strategy effectively reduces the number of nearby road boundaries that are grouped into a single curve.

\section{Evaluation with Real-World Driving Test}
\label{sec:evaluation}
% Dataset Description
We conducted a real-world driving test and collected a dataset comprising 30,424 frames of 4D mmWave radar point clouds from Changping District, China, for field evaluation. %Our 4DRadarRBD achieves accurate and robust performance on this dataset.

% Introduce the details about the dataset, what it include and how we do the labeling
\subsection{Driving Test and Dataset Description}
% Description of the dataset. Explain what it includes
The real-world driving dataset consists of 50 data clips, each approximately 40 seconds in duration, totaling 30,424 frames. During the driving tests, RGB cameras, 4D mmWave radar, LiDAR, GPS, and IMU sensors are used to capture detailed driving scenario information. RGB cameras and LiDAR capture detailed road scenarios for ground truth information. The GPS and IMU sensors provide the vehicle location, velocity, and yaw rate. For point-wise road boundary segmentation training, the training set includes 40 data clips, with 24,291 point cloud frames. To enhance model performance, data augmentation is applied by horizontally flipping the frames along the x-axis based on the left-right symmetry of the vehicle. With the data augmentation, the training set is expanded to 48,582 frames. Both the validation and test sets contain 5 data clips each, with 3,076 frames in the validation set and 3,057 frames in the test set. The dataset covers diverse driving scenarios, including highways, urban areas, and winding roads.
The ground truth labels of road boundaries are inferred from LiDAR point clouds using the PointPillars method~\cite{lang2019pointpillars}.

\subsection{Performance Evaluation Metrics}
% Describe evaluation metrics
We evaluate the 4DRadarRBD system using point-wise road boundary segmentation accuracy, Chamfer distance (CD) error, and Hausdorff distance (HD) error between the detected and actual road boundaries. Chamfer distance and Hausdorff distance are calculated as the average and maximum closest point distance, respectively, between the sets of detected and actual road boundary points~\cite{borgefors1986distance,rockafellar2009variational}.

\subsection{Evaluation Results and Ablation Study}

\begin{figure}[!t]
    \centering
    \includegraphics[width=0.49\textwidth]{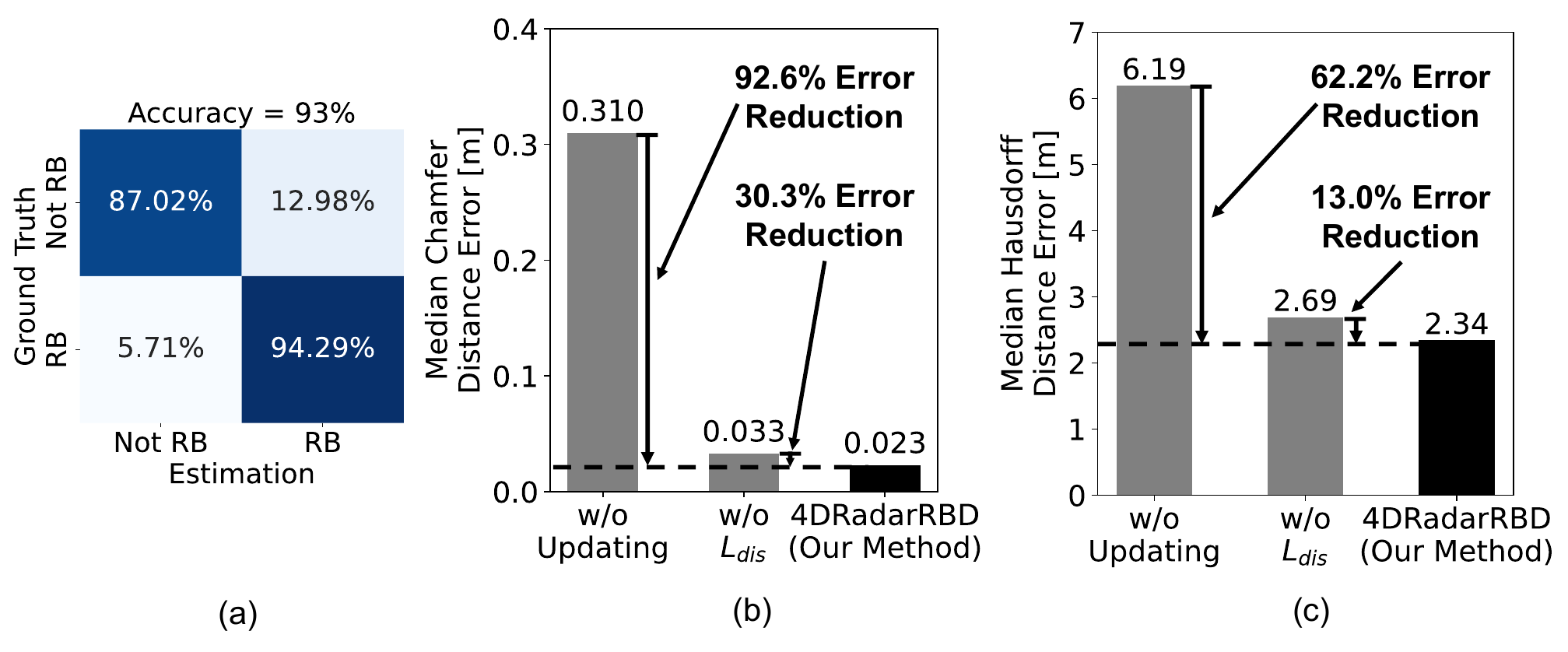}
    \caption{Overall Performance of 4DRadarRBD. (a) Confusion matrix for road boundary (RB) point segmentation (accuracy = 93$\%$), (b) Median Chamfer distance error of 4DRadarRBD (our method) and ablation tests without model updating and without distance loss, (c) Median Hausdorff distance error of 4DRadarRBD (our method) and ablation tests without model updating and without distance loss.}
    \label{fig:evaluation}
\end{figure}

4DRadarRBD achieves 93$\%$ accuracy for point-wise road boundary segmentation, with a median Chamfer distance error of 0.023 meters and a median Hausdorff distance error of 2.34 meters (see Fig.~\ref{fig:evaluation}). The confusion matrix for road boundary point segmentation is shown in Fig.~\ref{fig:evaluation} (a). Our method has high detection accuracy for both road boundary points and non-road boundary points.
% In most driving scenarios, such accuracies and errors can satisfy the requirements.

\begin{figure}[t]
    \centering
    \includegraphics[width=0.38\textwidth]{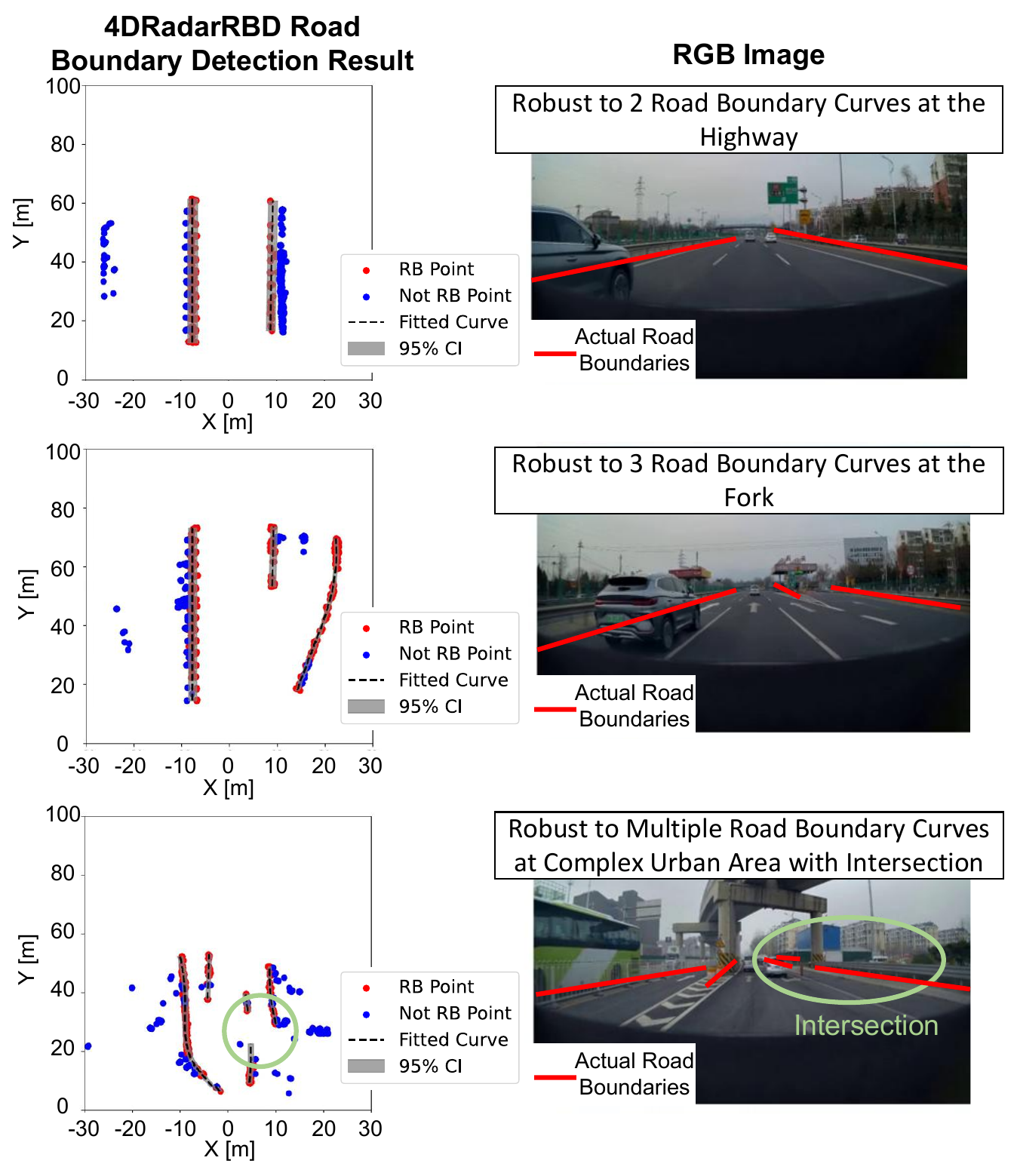}
    \caption{4DRadarRBD successfully detects road boundaries (referred to as RB) with varying numbers of road boundary curves in: a simple scenario with two curves (top), a forked road intersection with three curves (middle), and a complex urban environment with multiple curves (bottom). In the complex urban area, 4DRadarRBD successfully detects the intersection.}
    \label{fig:num_curves}
\end{figure}

To evaluate the effectiveness of 4DRadarRBD, two ablation tests are conducted including: 1) using the model without updating with the deviation vector obtained from the previous frame's detection results and 2) using the model trained without the incorporation of distance loss. By updating the model with the deviation vector, 4DRadarRBD reduces the median Chamfer distance error by 92.6$\%$ and the Hausdorff distance error by 62.2$\%$ (see Fig.~\ref{fig:evaluation} (b)). 4DRadarRBD reduces the median Chamfer distance error by 30.3$\%$ and the Hausdorff distance error by 13.0$\%$ by incorporating a distance loss to penalize false detections of points away from the road boundary (see Fig.~\ref{fig:evaluation} (c)). These results show that the proposed method of distance loss and updating the model with deviation vectors is effective.

\subsection{Evaluation of System Robustness}

In this section, we evaluate the robustness of the 4DRadarRBD system to varying numbers of road boundary curves, varying types of noisy points, varying road boundary curve shapes, and the system's temporal stability over driving time.

\subsubsection{Effect of Varying Numbers of Road Boundary Curves}

4DRadarRBD is robust to varying numbers of road boundary curves. 
Fig.~\ref{fig:num_curves} shows the road boundary detection results for three typical complex driving scenarios, including a highway situation with 2 road boundary curves (top), a fork road situation with 3 road boundary curves (middle), and a complex urban area situation with multiple road boundary curves (bottom). The left figures show the top view of the point cloud and the corresponding detection results, with the red dots representing the detected road boundary points and the blue dots representing the detected non-road boundary points. The right figures represent the corresponding RGB images. The 4DRadarRBD system can automatically detect varying numbers of road boundary curves and accurately fit the road boundary curves. Notably, in the third scenario, 4DRadarRBD successfully separates the two road boundary curves at the intersection instead of connecting them.

% \subsection{Performance on Various types of Roads (Straight / Curves)}

\begin{figure}[!t]
    \centering
    \includegraphics[width=0.38\textwidth]{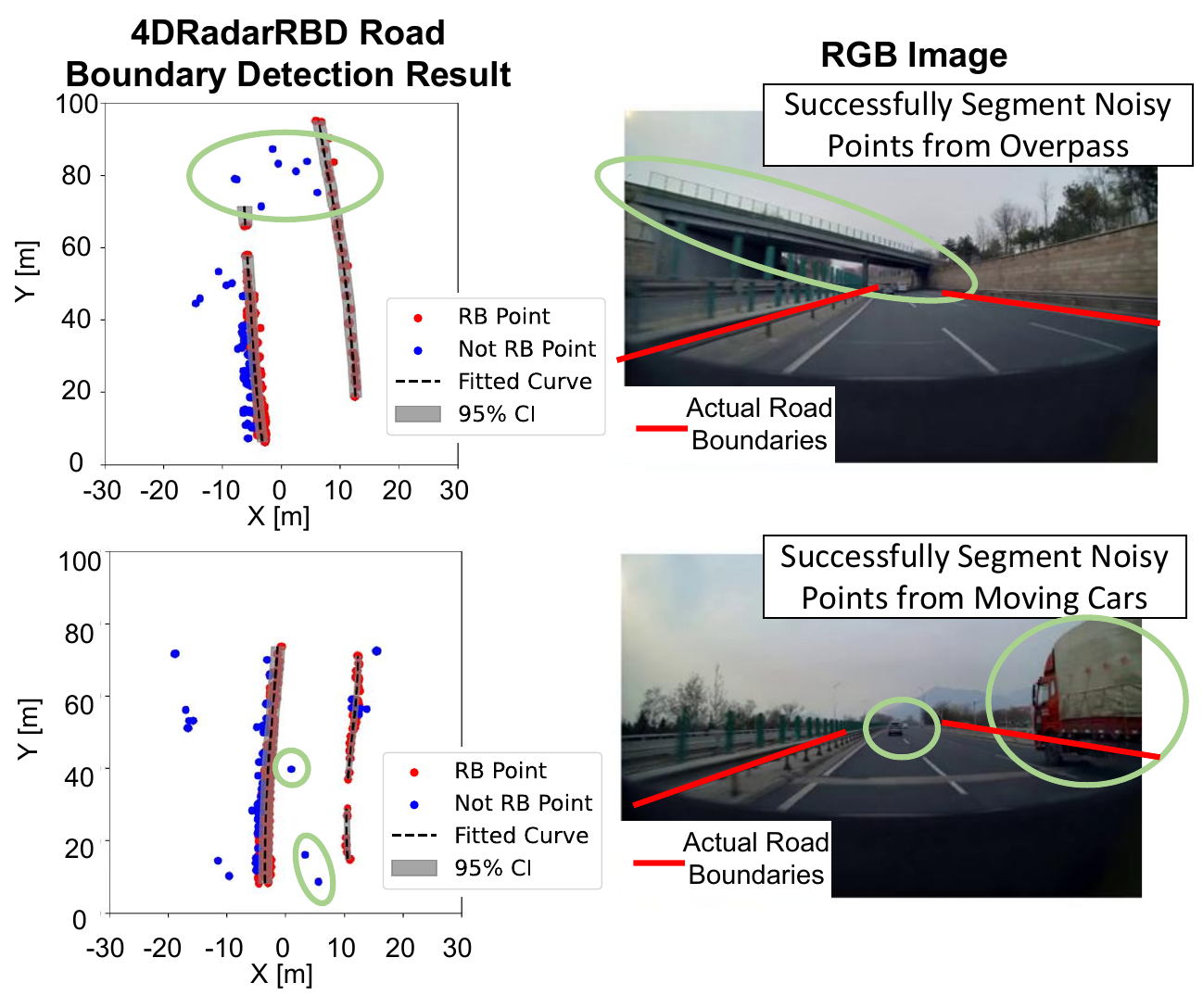}
    \caption{4DRadarRBD successfully segments road boundary (RB) points with the noisy points from overpasses and moving vehicles, proving its robustness in complex driving scenarios.}
    \label{fig:var_noise}
\end{figure}

\subsubsection{Effect of Noisy Points in Complex Driving Scenario}

The 4DRadarRBD system successfully achieves robust road boundary detection with various environmental noisy points. On a highway with multiple overpasses and moving vehicles, we achieve up to 94$\%$ accuracy for point-wise road boundary segmentation and a median Chamfer distance as low as 0.025m (see examples in Fig.~\ref{fig:var_noise}).

\subsubsection{Effect of Road Boundary Curve Shapes}
\begin{figure}[!t]
    \centering
    \includegraphics[width=0.38\textwidth]{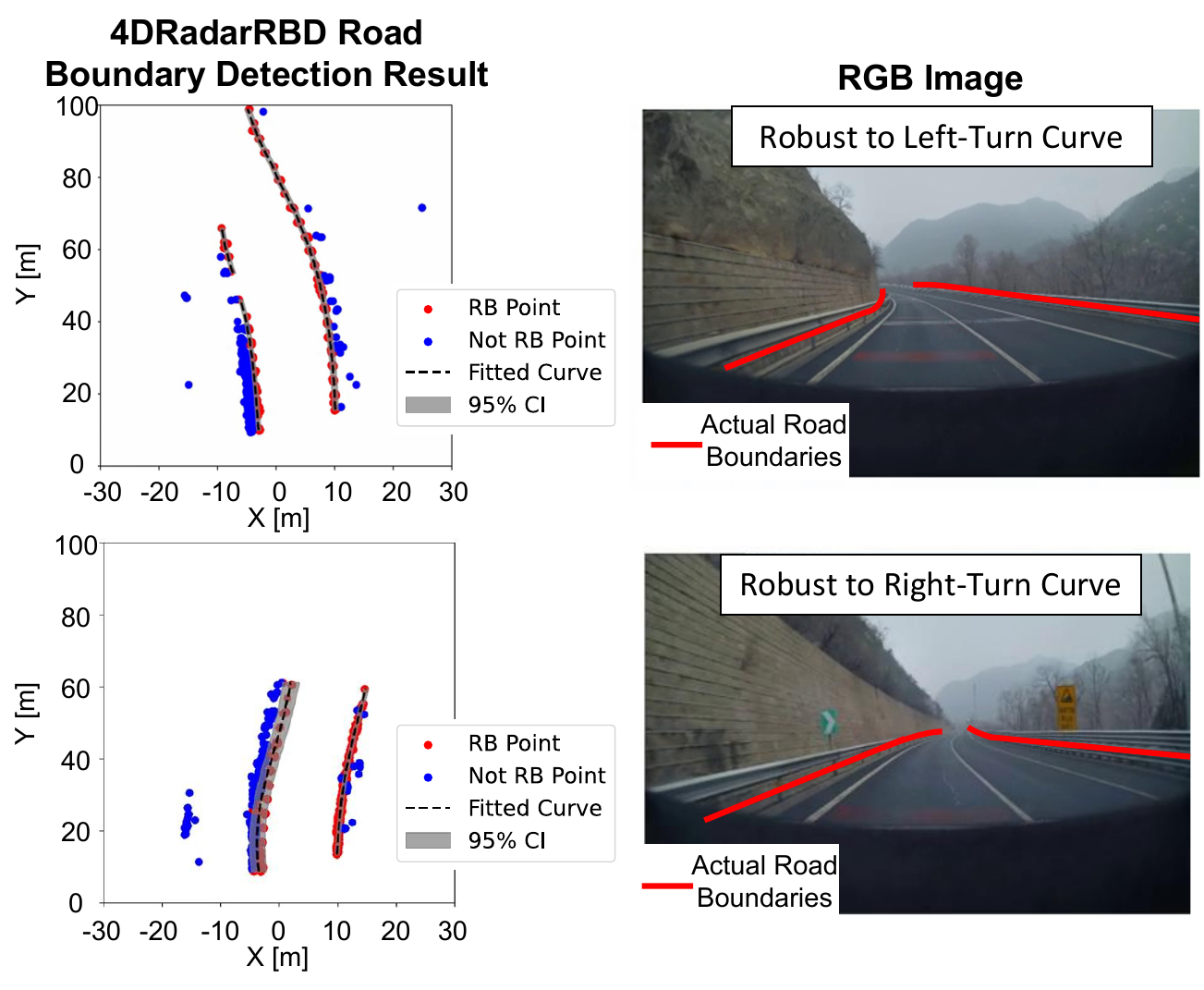}
    \caption{4DRadarRBD achieves robust detection performance for various road boundary curve shapes.}
    \label{fig:var_curve}
\end{figure}

4DRadarRBD is robust to various shapes of road boundary curves. Under the twisting driving conditions of mountainous roads, we achieve a point-wise road boundary segmentation accuracy as high as 91.2$\%$ and successfully fit various shapes of the road boundaries with a median Chamfer distance of 0.11m (see an example in Fig.~\ref{fig:var_curve}).

\begin{figure}[t]
    \centering
    \includegraphics[width=0.32\textwidth]{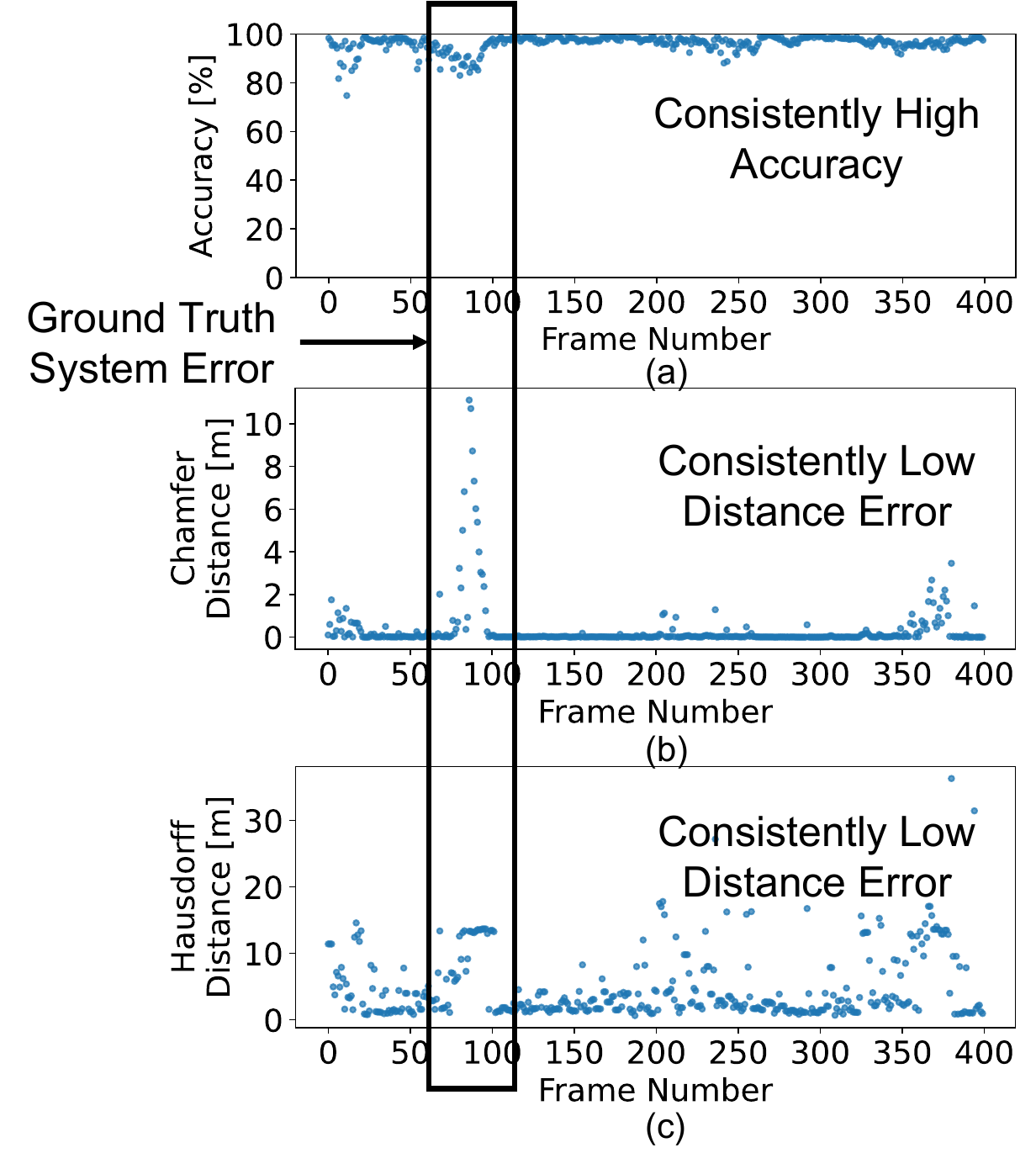}
    \caption{Consistently high segmentation accuracy, low Chamfer distance error, and low Hausdorff distance error for road boundary detection over time.}
    \label{fig:timehistory}
\end{figure}

\subsubsection{Sytem Temporal Stability}

Overall, 4DRadarRBD is robust with consistently high accuracy and low Chamfer and Hausdorff distance errors over continuous driving time (see Fig.~\ref{fig:timehistory}). At around frame 85, the decreased accuracy and increased distance errors are due to the ground truth system erroneously not detecting the right-hand side road boundary curve. Even in this case, our 4DRadarRBD successfully detects the road boundaries.

\section{Related Works}
\label{sec:related_work}
% Shorten this part
% How these are related to our system and method, this section can be shortened
% \subsection{Camera and LiDAR-based Road Boundary Detection Methods}

LiDAR and RGB cameras are mainly used for road boundary detection in autonomous driving. LiDAR systems provide high-resolution three-dimensional point clouds of the driving environment~\cite{medina2021machine,sun20193d}, which contain rich environmental information, enabling effective road boundary detection. RGB cameras are widely used for road boundary detection due to their affordability and wide availability in autonomous vehicles~\cite{chen2017rbnet, wen2008road, taher2018proposed}. Cameras capture extensive environmental details to identify lane markings, road edges, and barriers. However, both approaches have their limitations. LiDAR systems are costly, making them impractical for low-end vehicles. Cameras are vulnerable to poor lighting conditions. 

Previous studies have made preliminary attempts at 3D mmWave radar-based road boundary detection methods ~\cite{xu2020road, patel2022road, guo2014road, patel2024deep}. However, these methods lack robustness in complex road boundary geometries. This is because they usually utilize rule-based methods and algorithms such as RANSAC to fit road boundary curves, which are difficult to adapt to the dynamic and irregular road edge profiles common in complex environments due to the inherent limitations of polynomial models. To this end, there is a need for a more adaptive and robust approach that can automatically handle diverse road boundary geometries in real-world driving scenarios.

Point cloud segmentation methods fall into three categories: point-based~\cite{zhu2021adversarial, zhang2022towards, wu2019ground}, voxel-based~\cite{he2021vi, park2023pcscnet}, and projection-based~\cite{lang2019pointpillars, sun2024pepillar}. 
While voxel-based and projection-based methods are good at capturing spatial details, they are less effective for sparse 4D mmWave point clouds, often resulting in empty voxels or pixels in regions farther from the vehicle. Point-based methods, which directly process point clouds without altering data structure, are better suited for our road boundary segmentation task. Therefore, a point-based framework is used for road boundary segmentation.

\section{Conclusions}
\label{sec:conclusion}
% This part needs lots of modifications.
In this paper, we introduce 4DRadarRBD, the first 4D mmWave radar-based road boundary detection system that is cost-efficient and robust for complex driving scenarios. 
We reduce the noisy points by filtering via physical constraints and then segmenting the road boundary points with a distance-based loss. In addition, we capture the temporal dynamics of the point cloud using the vector representing the deviation of the current point from the motion-compensated road boundary detection result from the previous frame.
To evaluate 4DRadarRBD, we conducted a real-world driving test in Changping District, China. 
4DRadarRBD achieves accurate and robust road boundary detection in various complex driving scenarios with 93$\%$ accuracy for road boundary segmentation and a median Chamfer distance error of 0.023m.

\bibliographystyle{IEEEtran}
\bibliography{reference}

% \end{thebibliography}

\newpage

 

% \end{IEEEbiography}



\vfill

\end{document}